\documentclass{article}

\usepackage[preprint]{neurips_2026}
\usepackage[utf8]{inputenc}
\usepackage[T1]{fontenc}
\usepackage{microtype}
\usepackage{hyperref}
\usepackage{url}
\usepackage{booktabs}
\usepackage{amsfonts}
\usepackage{amsmath}
\usepackage{amssymb}
\usepackage{amsthm}
\usepackage{mathtools}
\usepackage{nicefrac}
\usepackage{graphicx}
\usepackage{subcaption}
\usepackage{xcolor}
\usepackage{multirow}
\usepackage{algorithm}
\usepackage{algorithmic}
\usepackage{comment}
\usepackage{tikz}
\usepackage{wrapfig}
\usepackage{comment}
\usepackage{siunitx}
\usetikzlibrary{arrows.meta,positioning,fit,calc,decorations.pathreplacing,patterns,backgrounds}

\newtheorem{definition}{Definition}

\newcommand{\cX}{\mathcal{X}}
\newcommand{\cM}{\mathcal{M}}

\newcommand{\cC}{\mathcal{C}}
\newcommand{\R}{\mathbb{R}}
\newcommand{\Enc}{E_\theta}
\newcommand{\proj}{\pi}
\newcommand{\joint}{\phi}
\newcommand{\muP}{\mu_P}
\newcommand{\muQ}{\mu_Q}
\newcommand{\SigP}{\Sigma_P}
\newcommand{\SigQ}{\Sigma_Q}

\newcommand{\Tr}{\operatorname{Tr}}

\newcommand{\Normal}{\mathcal{N}}

\newcommand{\MID}{\ifmmode\mathrm{MID}\else$\mathrm{MID}$\fi}

\newcommand{\todo}[1]{{\color{black}#1}}
\title{MID: Mask-Image Distributional Divergence for
       Evaluating Medical Image Segmentation}

\author{%
  \textbf{Vincenzo Marcian\`o}$^{1,2,3}$ \quad
  \textbf{Xiaoming Zhang}$^{1,2}$ \quad
  \textbf{Gianluca Guglielmo}$^{4}$ \\
  \textbf{Michela Antonelli}$^{3}$ \quad
  \textbf{S\'ebastien Ourselin}$^{3}$ \quad
  \textbf{Maria A Zuluaga}$^{1,3}$ \\[4pt]
  $^{1}$EURECOM \quad
  $^{2}$Sorbonne Universit\'e \quad
  $^{3}$King's College London \quad
  $^{4}$Technische Universit\"at Graz \\[2pt]
  {\small\texttt{\{vincenzo.marciano, xiaoming.zhang, maria.zuluaga\}@eurecom.fr}} \\
  {\small\texttt{\{michela.antonelli, sebastien.ourselin\}@kcl.ac.uk} \quad
  \texttt{gianluca.guglielmo@tugraz.at}}
}

\begin{document}

\maketitle

\begin{abstract}
Current evaluation protocols for medical image segmentation remain largely anchored in sample-wise comparisons, e.g., Dice and Hausdorff distance, which require each prediction to be paired with a ground-truth mask. Yet two limitations arise in practice: they are inapplicable when predictions are made on unlabeled data; and even when annotations are available, they cannot assess whether a collection of predictions matches the distributional structure of the ground-truth population.
We introduce \MID{} (Mask–Image Distributional Divergence), a distribution-level evaluation metric for medical image segmentation. \MID{} builds upon the observation that segmentation quality is a property of the joint image-mask distribution: a mask may be anatomically plausible in isolation yet incorrect for the image it accompanies. \MID{} compares the joint distribution of images and masks between a reference set and a set of predictions, without requiring sample-wise correspondence. This captures both mask realism and image-mask consistency. We evaluate \MID{} across 13 publicly available datasets spanning CT, MRI, and MRA. Under anatomically realistic corruptions across 56 organ groups, \MID{} tracks mean Dice and Hausdorff distance without per-sample ground-truth pairing. \MID{} also detects distributional shifts across modalities and anatomies, identifies image-mask misalignment through controlled shuffling experiments, and produces reliable model rankings across real segmentation models. Official implementation available at \texttt{\href{https://github.com/erc-caravel/pytorch-mid}{github.com/erc-caravel/pytorch-mid}}

\end{abstract}

\section{Introduction}
\label{sec:intro}

Deep learning has substantially advanced automatic medical image segmentation, yielding models that range from single-anatomy specialists~\cite{isensee2021nnu,myronenko20183d,liu2021swin} to pan-organ foundation models~\cite{isensee2025nninteractive,MedSAM,ma2025medsam2,wang2023sammed3d,wong2024scribbleprompt}. Evaluating these models 
remains fundamentally tied to per-sample ground-truth (GT) annotations at test time.
Dice similarity coefficient, Hausdorff distance (HD95), and normalised surface distance (NSD)~\cite{maier2024metrics} each measure how accurately a predicted mask reproduces the paired GT annotation at the voxel level. Although they were  effective in controlled experimental conditions, several practical challenges remain to be addressed. First, every prediction must be matched to an annotated GT, an assumption that fails during prospective deployment and when evaluating foundation models on unlabeled cohorts. Second, when GT annotations are unavailable, an alternative is to generate surrogate references using encoder-decoder refinement models~\cite{cosarinsky2026rca,galati2021efficient,larrazabal2020post,painchaud2020,wang2020deep}; however, these approaches require training additional models for each target dataset, which is computationally expensive and limits scalability.

Perceptual image quality assessment has similarly moved beyond pixel-level comparison toward feature-level distances in pretrained encoder spaces. Learned Perceptual Image Patch Similarity (LPIPS)~\cite{zhang2018unreasonable} showed that $\ell_2$ distances between VGG features of two images correlate with human perceptual judgements far better than Structural Similarity Index Measure (SSIM) or pixel norms, because pretrained encoders capture semantic structure rather than low-level coincidences. This idea extends naturally from per-sample distances to population-level comparisons.
Once samples are embedded in a feature space, a dataset induces an empirical distribution, and the squared \num{2}-Wasserstein distance $W_2^2$ provides a principled way to compare such distributions: it respects the geometry of the embedding space and admits a closed-form expression for Gaussian distributions~\cite{dowson1982frechet}.  The Fr\'{e}chet Inception Distance (FID) applied this principle to generative image evaluation: images are embedded into Inception-v3 features, Gaussian distributions are fitted to the reference and generated sets, and the squared Fr\'{e}chet distance, equivalent to $W_2^2$ between the two fitted Gaussians, is reported as a single scalar. FID requires no per-sample correspondence between reference and candidate sets, correlates with human judgements of image quality, and has become the de facto standard in generative modelling.

Feature-space distribution matching therefore offers a promising route toward segmentation evaluation without per-sample ground truth. However, extending this paradigm from image generation to medical image segmentation is non-trivial.
First, segmentation quality is inherently defined over the \emph{joint} distribution of images and masks: a mask may be anatomically plausible in isolation, yet clinically incorrect if it is misaligned with the anatomy in the corresponding image. Image-only distributional metrics cannot capture such failures. Second, medical imaging spans both volumetric and 2-D modalities (e.g., CT, MRI). A general-purpose metric should  compare image-mask distributions across these settings without relying on modality-specific pipelines.

We propose \MID{} (Mask-Image Distributional Divergence),  a distribution-level metric for medical image segmentation that shifts evaluation from sample-wise mask comparison to probability-measure comparison over the joint space $\cX \times \cM$. Specifically, \MID{} maps each image and its class-specific mask to a joint representation through a pretrained medical vision encoder, thereby inducing two empirical push-forward distributions: one from ground-truth image-mask pairs and one from predicted image-mask pairs. \MID{} then fits Gaussian approximations to these embedded distributions and reports their squared \num{2}-Wasserstein distance. By comparing distributions rather than aligned samples, \MID{} requires no per-case correspondence between reference and candidate sets, while remaining sensitive to the image-mask dependence that defines segmentation quality.

This construction adapts feature-space distribution matching to the structured image-mask dependency inherent in segmentation. \MID{} captures failure modes invisible to single-modality metrics: unlike image-only approaches, it detects mask-image inconsistency; unlike mask-only metrics, it identifies anatomically implausible placements even when mask shapes are realistic.

Across \num{56} organ groups, using five signed-distance-transform (SDT)-based corruption types at three severity levels, \MID{} increases consistently as segmentation quality degrades. Without requiring any per-sample correspondence, \MID{} correlates with mean Dice at Spearman $\rho=-0.55$. For comparison, HD95, a paired-reference geometric metric, reaches $|\rho|=0.72$ with Dice on the same corrupted masks but requires a ground-truth mask for every prediction.  While per-sample metrics remain more precise when ground truth is available, \MID{} provides a complementary distribution-level signal for settings where candidate predictions cannot be annotated case by case.
%

Our contributions are as follows: 
(1) we formulate segmentation evaluation without paired ground truth as distribution matching on the joint image-mask space and introduce \MID{}; 
(2) we propose a diagnostic decomposition of \MID{} into image, mask, and residual correspondence terms; 
and (3) across thirteen datasets, we show that \MID{} tracks corruption severity, detects distribution shifts, identifies image-mask misalignment, and ranks models consistently with paired metrics.


\section{Related Work}
\label{sec:related}

\paragraph{Per-sample segmentation metrics.}
The Metrics Reloaded framework \cite{maier2024metrics} catalogues overlap measures (Dice, IoU), surface distances (HD95, NSD, ASSD), and detection metrics. Topology-preserving variants include clDice \cite{shit2021cldice} and topologically faithful segmentation \cite{stucki2023topologically}. All of these metrics require per-sample GT alignment at test-time.

\paragraph{FID and distribution-level evaluation.}
FID \cite{heusel2017gans} embeds images into Inception-v3 features and computes the squared Fr\'{e}chet distance \cite{dowson1982frechet}, which equals $W_2^2$ between the fitted Gaussians \cite{Courty2017JointDO}. It correlates with human quality judgements at a scale that per-image metrics cannot. Kernel Inception Distance (KID) \cite{binkowski2018demystifying} provides an unbiased Maximum Mean Discrepancy (MMD)-based alternative. FID has been extended to joint image-caption embeddings \cite{devries2019evaluation} by concatenating image and conditioning features before fitting the Gaussian; however, this setting is limited to 2-D natural images and does not model structured segmentation masks or volumetric data. \MID{} addresses these limitations by jointly encoding images and masks, supporting both 2-D and 3-D inputs, and enabling a decomposition into image, mask, and residual correspondence terms.


\paragraph{Medical imaging representation learning.}
Recent work shows that learned feature spaces for medical images organize by anatomy and generalize across datasets and modalities~\cite{pinaya2022brain,perez2025exploring}. Early self-supervised and generative approaches demonstrated this structure without label supervision, with masked autoencoders and latent diffusion models recovering clinically relevant representations in brain MRI and abdominal CT~\cite{bercea2023mask,guo2025maisi,pinaya2022brain}. Scaling vision-only models, such as Rad-DINO~\cite{perez2025exploring}, and contrastive image-text pretraining, including BiomedCLIP~\cite{zhang2023biomedclip} and UniMedCLIP~\cite{khattak2024unimed}, further improved generalisation across anatomies, scanner types, and modalities. More recent large-scale medical vision encoders~\cite{sellergren2025medgemma,zhai2023sigmoid} produce shared feature spaces spanning CT, MRI, pathology, and fundus photography. Feature-space distances from such encoders are widely used for out-of-distribution detection, e.g., via Mahalanobis distance~\cite{lee2018simple}. However, these approaches operate on images alone and do not model image-mask correspondence, a key aspect of segmentation evaluation.

\section{Method}
\label{sec:method}

\MID{} evaluates segmentation quality through the joint image-mask distribution rather than image or mask marginals alone. It is constructed in three steps. First, for each image-mask pair, the image is encoded once, and each foreground class is encoded as a separate binary mask
using the same medical vision encoder for both the image and mask inputs, producing one joint vector per foreground class. Second, the resulting vectors are projected to a lower-dimensional space via PCA fitted on the reference set. Finally, the squared 2-Wasserstein distance between Gaussian approximations of the reference and candidate distributions is computed.
\todo{Sections~\ref{sec:notation}--\ref{sec:decomposition} formalise these steps.}

\subsection{Setup and Joint Embedding}
\label{sec:notation}

Let $\cX$ denote the space of 2-D and 3-D medical images, and let $\cM$ denote the space of $C$-class segmentation masks defined on the same grid.
For a segmentation mask $m \in \cM$, let $\cC(m) \subseteq \{1,\ldots,C\}$ denote the set of foreground classes present in $m$. For each $c \in \cC(m)$, let $m_c=\mathbf{1}[m=c]$ be the corresponding binary mask, so that $m$ is represented as the collection of class-specific binary masks $\{m_c\}_{c\in\cC(m)}$.


Let $f_\theta$ denote a medical vision encoder that maps an input image or mask rendering to a $d$-dimensional feature vector in $\R^d$. For image inputs, $f_\theta$ produces vectors in the same shared feature space, allowing 2-D and 3-D samples to be represented within a single joint distribution.

\begin{definition}[Per-class joint feature embedding]
\label{def:joint}
For an image-mask pair $(X, m) \in \cX \times \cM$ and foreground class $c\in\cC(m)$, the per-class joint feature embedding is defined as
%
\begin{equation}
  \joint_c(X, m)
  = \proj\!\left(\bigl[f_{\theta}(X) \;\|\; f_{\theta}(m_c)\bigr]\right)
  \in \R^k,
  \label{eq:joint_embedding}
\end{equation}
where $[\cdot \| \cdot]$ denotes the concatenation of the image and mask embeddings and $\proj : \R^{2d} \to \R^k$ is a projection. 
\end{definition}

Each $\joint_c(X,m)$ represents the pairing between image $X$ and the class-specific binary mask $m_c$. For a dataset $\{(X_i,m_i)\}_{i=1}^N$, each image-mask pair contributes one such embedding for every foreground class present in the mask. The resulting collection $\{\joint_c(X_i,m_i)\}_{i=1,\ldots,N;\,c\in\cC(m_i)}$
forms the embedding set over which \MID{} fits its Gaussian.

In practice, the image embedding is computed once per volume and reused across foreground classes. Class masks are encoded as binary masks with a shared foreground label, so their embeddings capture geometry rather than class identity. The embedding is deterministic, ensuring reproducibility without requiring fixed random seeds.

\subsection{MID}
\label{sec:MID}
\label{sec:theory}

Let $P$ be a reference distribution over ground-truth image-mask pairs and $Q$ a candidate distribution over predicted pairs. Following Section~\ref{sec:notation}, each image-mask pair contributes one embedding per foreground class present in the mask. The empirical distributions for $P$ and $Q$ are obtained by pooling these embeddings before fitting Gaussian approximations.
%
Let $(\muP, \SigP)$ and $(\muQ, \SigQ)$ be the empirical means and covariances of the projected embeddings $\{\joint_c(X_i, m_i)\}_{i=1,\ldots,N; \,c\in\cC(m_i)}$ 
under $P$ and $Q$.

\begin{definition}[\MID{}]
\label{def:MID}
\begin{equation}
\begin{aligned}
\text{MID}(P, Q) 
&= W_2^2\!\bigl(\Normal(\muP,\SigP),\,\Normal(\muQ,\SigQ)\bigr) \\
&= \|\muP - \muQ\|^2 \\
&\quad + \Tr\!\Bigl(\SigP + \SigQ
- 2 \bigl(\SigP^{1/2} \SigQ \SigP^{1/2}\bigr)^{1/2}\Bigr).
\end{aligned}
\label{eq:MID}
\end{equation}
\end{definition}

The closed-form expression in Equation~\eqref{eq:MID} follows from Dowson \& Landau \cite{dowson1982frechet} and corresponds to the squared 2-Wasserstein distance between Gaussian approximations of the two embedding distributions, as used in Fréchet Inception Distance (FID)~\cite{heusel2017gans}.
$P$ and $Q$ need not be aligned sample by sample, as \MID{} compares distributions rather than individual pairs.



\subsection{Projection Design}
\label{sec:projection}
The joint embedding has dimension $2d$, which is high relative to the size of typical medical imaging datasets. In this setting, reliable covariance estimation is challenging and requires dimensionality reduction. To this end, the embeddings are projected to a lower-dimensional space using PCA fitted on the reference set. This projection preserves the dominant directions of variation while improving the stability of the covariance estimates required for \MID{}.
In practice, we use $k = 128$ components, which retain more than \SI{94}{\percent} of the variance in the reference set.
Under a Gaussian assumption, the resulting approximation preserves \MID{} up to the discarded variance, with $|\MID_k - \MID| \leq 2\sum_{i>k}\lambda_i$, where $\lambda_i$ are the eigenvalues of the covariance. 

\subsection{Sensitivity Decomposition}
\label{sec:decomposition}
The joint \MID{} score can be further analysed by separating image appearance, mask geometry, and image-mask correspondence. We compute two auxiliary Fréchet distances using the same estimator as in Equation~\eqref{eq:MID}: an image-only term $\MID_{\mathrm{img}}$ computed from image embeddings $f_\theta(X)$, and a mask-only term $\MID_{\mathrm{mask}}$ computed from mask embeddings $f_\theta(m_c)$. Concretely, $\MID(P,Q)$ is computed on joint embeddings, $\MID_{\mathrm{img}}(P,Q)$ on image-only embeddings, and $\MID_{\mathrm{mask}}(P,Q)$ on mask-only embeddings, each using the same Gaussian Fréchet estimator and reference-fitted projection protocol. The remaining contribution is attributed to image-mask correspondence and defines the residual correspondence term:
\begin{equation}
  \MID_{\mathrm{corr}}(P,Q)
  \;:=\;
  \MID(P,Q) - \MID_{\mathrm{img}}(P,Q) - \MID_{\mathrm{mask}}(P,Q).
  \label{eq:decomp}
\end{equation}

Unlike $\MID_{\mathrm{img}}$ and $\MID_{\mathrm{mask}}$, $\MID_{\mathrm{corr}}$ is a diagnostic residual rather than a standalone distance. It captures the part of the joint score not explained by image-only or mask-only shifts. Perturbations that preserve both image-only and mask-only distributions but alter their dependence structure, such as random image-mask shuffling, leave $\MID_{\mathrm{img}}$ and $\MID_{\mathrm{mask}}$ approximately unchanged while increasing $\MID_{\mathrm{corr}}$. The derivation of Equation~\ref{eq:decomp} is given in Appendix~\ref{app:cov_block}.

\section{Experiments}
\label{sec:experiments}

We evaluate \MID{} through a series of controlled and real-world experiments. 
Experiment 1 assesses sensitivity to controlled, anatomically realistic segmentation corruptions. 
Experiment 2 examines cross-dataset distributional shift. 
Experiment 3 compares distributional metrics against per-sample geometric metrics. 
Experiment 4 evaluates \MID{} on outputs from real segmentation models.
Experiment 5 isolates the role of image-mask correspondence via controlled perturbations. 
We also provide additional analyses on within-group spread (Appendix~\ref{app:exp1_groups}), sample complexity (Appendix~\ref{app:sample_complexity}), dataset-specific projections (Appendix~\ref{app:exp8}), and latent space geometry (Appendix~\ref{app:latent_vis}).


\subsection{Setup}
\label{sec:setup}

\noindent\textbf{Datasets.}
We use thirteen publicly available datasets spanning CT, MRI, and magnetic resonance angiography (MRA): AbdomenCT-1K~\cite{ma2021abdomenct}, ACDC~\cite{bernard2018deep}, AMOS CT and MRI~\cite{ji2022amos}, IXI MRA with VesselVerse annotations~\cite{ixi2006dataset,falcetta2025}, six Medical Segmentation Decathlon tasks (Task03 Liver, Task05 Prostate, Task06 Lung, Task07 Pancreas, Task09 Spleen, Task10 Colon)~\cite{antonelli2022medical}, and TotalSegmentator v2 CT and MRI. Dataset-specific foreground classes and sample counts are reported in Appendix~\ref{app:datasets}.


\noindent\textbf{Reference distribution.}
\todo{Throughout the paper we count multimodal datasets by modality (so AMOS contributes AMOS\,CT and AMOS\,MRI, and TotalSegmentator v2 contributes TotalSeg\,CT and TotalSeg\,MRI), giving the \num{13} datasets listed above. Unless otherwise stated, the reference distribution is built from clean image--ground-truth mask pairs in the reference split of a \num{10}-dataset reference pool: ACDC, IXI~MRA (VesselVerse), the six MSD tasks (Task03 Liver, Task05 Prostate, Task06 Lung, Task07 Pancreas, Task09 Spleen, Task10 Colon), and TotalSegmentator~v2 CT and MRI. The remaining three datasets, AbdomenCT-1K, AMOS\,CT, and AMOS\,MRI, are excluded from the reference pool and reserved as held-out candidate sets for cross-dataset evaluation.}

\noindent\textbf{Candidate distributions.}
Candidate distributions are experiment-specific and are compared against the fixed reference distribution. We construct candidate distributions from two sources. First, controlled synthetic corruptions are applied to ground-truth masks to produce candidate masks with known degradation type and severity. Following controlled experimental designs in~\cite{arestafaz,jebril2025shape}, we use five signed-distance-transform (SDT) based corruption types: elastic boundary deformation, smooth erosion, smooth dilation, blob-shaped false positives, and partial missing slices, each at mild, moderate, and severe levels (Appendix~\ref{app:corruptions}). Second, we use predictions from real segmentation models. SwinUNETR~\cite{liu2021swin} and MedSAM~\cite{ma2025medsam2} are kept frozen and used directly for inference: SwinUNETR is applied to abdominal CT datasets, while MedSAM is applied across non-vascular datasets using synthetic prompts derived from ground-truth masks. SegResNet~\cite{myronenko20183d} is trained on all non-vascular datasets and evaluated at \num{50}, \num{150}, and \num{300} epochs, yielding predictions at increasing quality levels. For vascular segmentation on IXI MRA, we use model predictions released with VesselVerse: nnUNet~\cite{isensee2021nnu}, A2V~\cite{galati2023a2v}, and SPOCKMIP~\cite{radhakrishna2024spockmip}.


\noindent\textbf{Baselines.}
We compare \MID{} against five baselines. \textbf{FID-Inception} computes standard FID using Inception-v3~\cite{heusel2017gans} features on axial mid-slice renderings; this 2-D proxy discards volumetric information and mask information. \textbf{FID-img} computes the Fréchet distance on image features from $f_\theta$ only, without masks. \textbf{FID-mask} computes the Fréchet distance on mask features from $f_\theta$ only, without images. \textbf{Mahalanobis} computes the mean squared Mahalanobis distance from a reference Gaussian in the joint feature space. \textbf{KID (Kernel Inception Distance)}~\cite{binkowski2018demystifying} is an unbiased distributional baseline computed on mask features. Finally, \textbf{per-sample geometric metrics}, mean Dice and HD95 over matched  --ground-truth pairs, are used as reference.

\noindent\textbf{Implementation.} 
For the main experiments, we instantiate $f_\theta$ with MedSigLIP~\cite{sellergren2025medgemma}. For 3-D volumes, we follow the original MedSigLIP protocol and average patch-token embeddings across axial slices; for MAISI-based encoders (Appendix~\ref{app:encoder_choice}), each 2-D axial slice is encoded as a 3-D input with depth 1. Alternative encoder choices are evaluated in Appendix~\ref{app:encoder_choice}. We pre-extract MedSigLIP features for all datasets, yielding \todo{\num{17631}} per-class image-mask feature representations \todo{(of which \num{11223} belong to the \num{10}-dataset reference pool and \num{6408} to the held-out AMOS\,CT, AMOS\,MRI, and AbdomenCT-1K candidate sets).} Unless otherwise stated, \MID{} and the distribution-based baselines are computed using the same encoder $f_\theta$, with feature inputs selected according to each metric. We use incremental PCA with $k=128$ components, fitted on the reference set. Covariance is estimated via Ledoit-Wolf shrinkage~\cite{ledoit2004well}. All code was implemented in Python 3.10 using PyTorch 2.0 and CUDA 12.4. All experiments use a fixed random seed. Code is available at \texttt{[anonymous]}.

\subsection{Response to Controlled Segmentation Corruptions}
\label{sec:exp1}
Experiment 1 assesses whether \MID{} responds monotonically to controlled, anatomically realistic mask corruptions of known severity. For each combination of dataset, foreground class, corruption type, and severity level, the candidate distribution is built from the evaluation split by replacing ground-truth masks with corrupted masks. \MID{} is computed against the global reference distribution defined in Section~\ref{sec:setup}, without sample-wise pairing between reference and candidate embeddings. Dice and HD95 are computed on the same evaluation set and serve as paired reference-quality indicators. The quantitative analysis is performed over all such combinations in the full corruption sweep. 

\begin{figure}[t]
\centering
\includegraphics[width=0.9\linewidth]{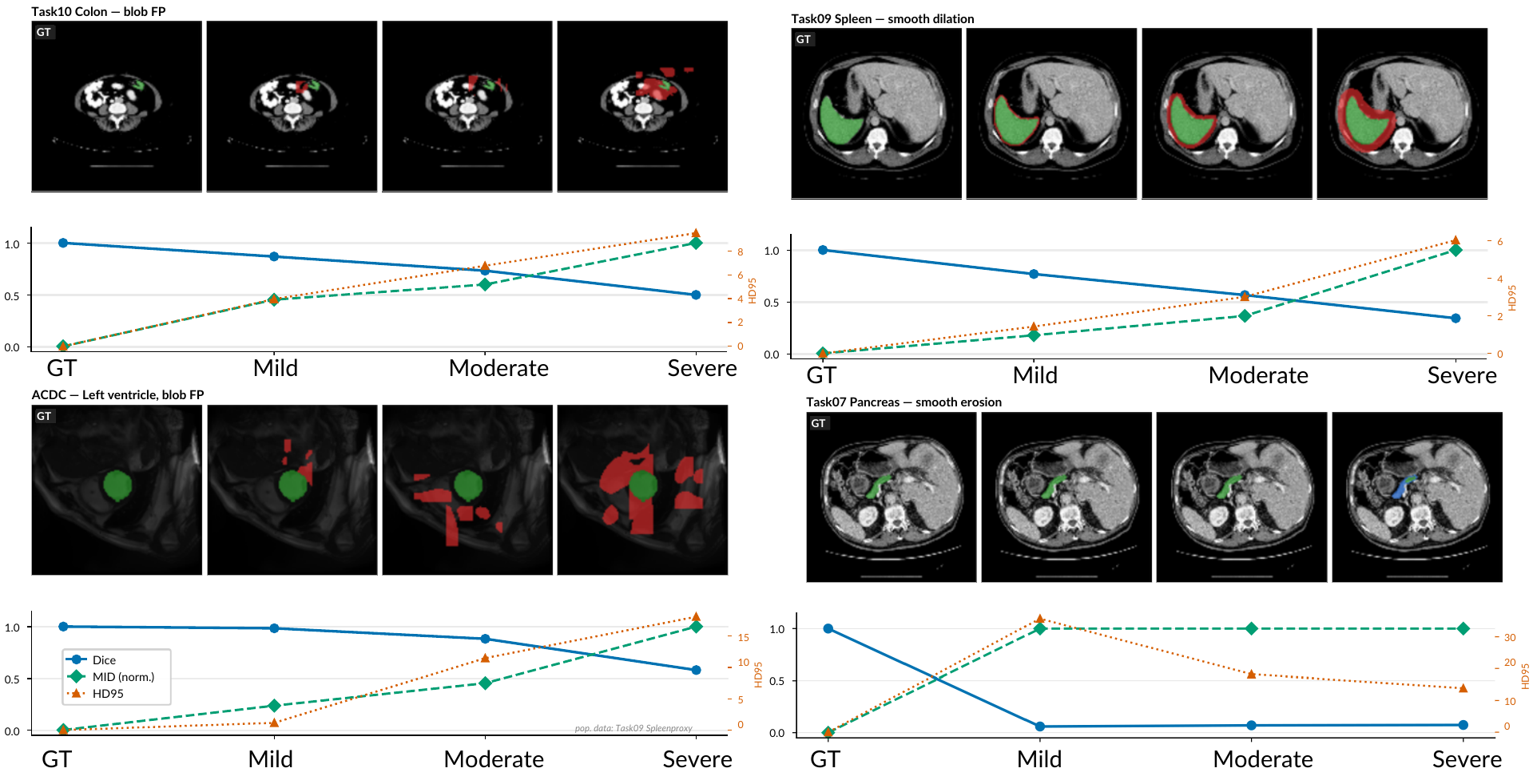}
\caption{
Representative examples from the controlled corruption sweep (Experiment~1). The four panels show Task10 Colon with blob-shaped false positives, Task09 Spleen with smooth dilation, ACDC left ventricle with blob-shaped false positives, and Task07 Pancreas with smooth erosion. Overlays show the ground-truth mask and corrupted masks at mild, moderate, and severe levels. Line plots show \MID{}, Dice, and HD95 as a function of degradation severity. \MID{} is normalised by the panel-wise maximum for visualisation only; all reported correlations use raw values.}


\label{fig:sensitivity}
\end{figure}

\todo{Figure~\ref{fig:sensitivity} shows four representative cases from the sweep, selected to span different datasets, anatomies, and degradation patterns. In each panel, \MID{} rises with degradation severity in step with the decline in Dice and the increase in HD95, mirroring per-sample paired metrics without using per-case ground-truth at inference time. \MID{} is normalised by the panel-wise maximum for visualisation only; all reported correlations use raw values.} \todo{Across the full $840$-cell sweep, \MID{} tracks both paired-reference indicators in the expected direction: it is negatively correlated with mean Dice (Spearman $\rho = -0.55$, $p < 0.001$), since lower \MID{} and higher Dice indicate better segmentations, and positively correlated with HD95 ($\rho = 0.72$, $p < 0.001$). Within each foreground class, \MID{} increases monotonically with corruption severity across all five corruption types.} \todo{Disaggregating by corruption type provides direct evidence that shape-altering corruptions yield stronger correlations with Dice than additive false-positive or missing-slice errors (Table~\ref{tab:per_corruption}). Smooth erosion, smooth dilation, and elastic boundary deformation reach $\rho$ between $-0.58$ and $-0.65$, whereas blob false positives and partial missing slices reach $-0.45$ and $-0.42$. This indicates that \MID{} responds not only to voxel-level disagreement but also to the type of anatomical perturbation.}

\todo{\begin{table}[t]
\centering
\begin{minipage}[b]{0.49\linewidth}
\centering
\captionof{table}{\MID{}--Dice Spearman correlation disaggregated by corruption type, over \num{56} organ groups $\times$ \num{3} severities (\num{168} cells per row). \SI{95}{\percent} bootstrap CIs use \num{1000} replicates.}
\label{tab:per_corruption}
\resizebox{\linewidth}{!}{%
\begin{tabular}{lcc}
\toprule
Corruption type & $\rho$ vs.\ Dice & \SI{95}{\percent} CI \\
\midrule
Smooth erosion          & $-0.65$ & $[-0.71,\,-0.58]$ \\
Smooth dilation         & $-0.62$ & $[-0.68,\,-0.55]$ \\
Elastic deformation     & $-0.58$ & $[-0.64,\,-0.51]$ \\
Blob false positives    & $-0.45$ & $[-0.52,\,-0.38]$ \\
Partial missing slices  & $-0.42$ & $[-0.49,\,-0.34]$ \\
\midrule
All (840 cells)         & $-0.55$ & $[-0.61,\,-0.49]$ \\
\bottomrule
\end{tabular}%
}
\end{minipage}\hfill
\begin{minipage}[b]{0.49\linewidth}
\centering
\captionof{table}{\MID{} between TotalSeg CT train (reference, \num{1105} cases, \num{6630} vectors) and candidate datasets, ordered by score. Mahalanobis distance (Mah.) shown for comparison.}
\label{tab:ood}
\resizebox{\linewidth}{!}{%
\begin{tabular}{lcrr}
\toprule
Candidate & Mod. & \MID{} & Mah. \\
\midrule
TotalSeg CT (held-out) & CT  & $0.011$ &  $11.2$ \\
TotalSeg MRI           & MRI & $0.278$ &  $34.7$ \\
Prostate MRI           & MRI & $0.932$ &  $48.8$ \\
ACDC cardiac           & MRI & $0.949$ &  $46.6$ \\
\bottomrule
\end{tabular}%
}
\end{minipage}
\end{table}}

\todo{The corruption sweep also reveals that the scale of \MID{} varies substantially across anatomies, spanning roughly an order of magnitude between compact structures (e.g., adrenal glands, vessels) and large organs (e.g., liver, lung lobes). As a result, absolute \MID{} values should be interpreted relative to anatomy-specific reference ranges, rather than compared directly across anatomies. Additional analyses, including the per-anatomy reference index (Appendix~\ref{app:atlas}, Figure~\ref{fig:atlas}), clean-reference variability (Appendix~\ref{app:exp1_groups}), and sample complexity (Appendix~\ref{app:sample_complexity}), are reported in the appendix.}

\subsection{Distributional Shift Detection}
\label{sec:exp2}
\MID{} compares a candidate distribution of image-mask pairs to a fixed reference distribution. It should therefore increase when the candidate set differs systematically from the reference, whether through changes in anatomy, modality, acquisition quality, or predicted mask geometry. In Experiment 2, we test two complementary modes of such distributional shift: cross-dataset shifts in anatomy or modality, and within-dataset shifts induced by image-acquisition artifacts.

\noindent\textbf{Cross-dataset shift.}
We fix a single reference distribution from TotalSegmentator CT train (\num{1105} cases, \num{6630} latent vectors across six abdominal organs: spleen, left kidney, right kidney, liver, stomach, and pancreas) and measure \MID{} against four candidate sets spanning in-distribution, modality-shift, and anatomy-shift settings, with the corresponding scores reported in \todo{Table~\ref{tab:ood}}. The in-distribution control is the TotalSegmentator CT held-out test split (\num{123} cases, \num{738} vectors), drawn from the same dataset and acquisition protocol as the reference. The resulting \MID{} $= 0.011$ provides an empirical finite-sample floor: even a held-out split from the same distribution yields a small non-zero score. All shifted candidates are interpreted relative to this baseline.
TotalSegmentator MRI increases \MID{} to \num{0.278}, a \num{23}$\times$ increase over the floor, reflecting a modality shift under matched abdominal anatomy. Prostate MRI and ACDC cardiac MRI produce larger shifts (\MID{} $=0.932$ and $0.949$), approximately \num{85}$\times$ above the floor, as anatomy, modality, and acquisition protocol all differ from the reference. This graded response shows that \MID{} is sensitive to the type of shift and not only to its presence.

\noindent\textbf{Within-dataset shift via image artefacts.}
We next apply ROOD-MRI~\cite{boone2023rood}, a robustness benchmark for controlled MRI acquisition artefacts, to Task05\_Prostate. Three artefact types (Ghosting, Rician noise, Random Motion) are applied at three severity levels (mild, moderate, severe). The reference distribution is pre-computed from clean T2 images with ground-truth masks; candidate distributions are built from corrupted images paired with SegResNet-300 predictions, without using ground-truth masks at inference time. Additional setup details are provided in Appendix~\ref{app:rood_prostate}.

Figure~\ref{fig:rood_prostate} shows that \MID{} tracks the degradation monotonically across all three artefact types and all severity levels, mirroring the Dice and HD95 trends without requiring any per-case ground truth at inference time. This setting differs from the controlled corruption experiment (Section~\ref{sec:exp1}), where image quality is fixed and only masks are corrupted. Here, both image appearance and predicted masks shift with artefact severity, and \MID{} captures the combined distributional change as a single divergence from the clean reference.

\begin{figure}[t]
\centering
\includegraphics[width=\linewidth]{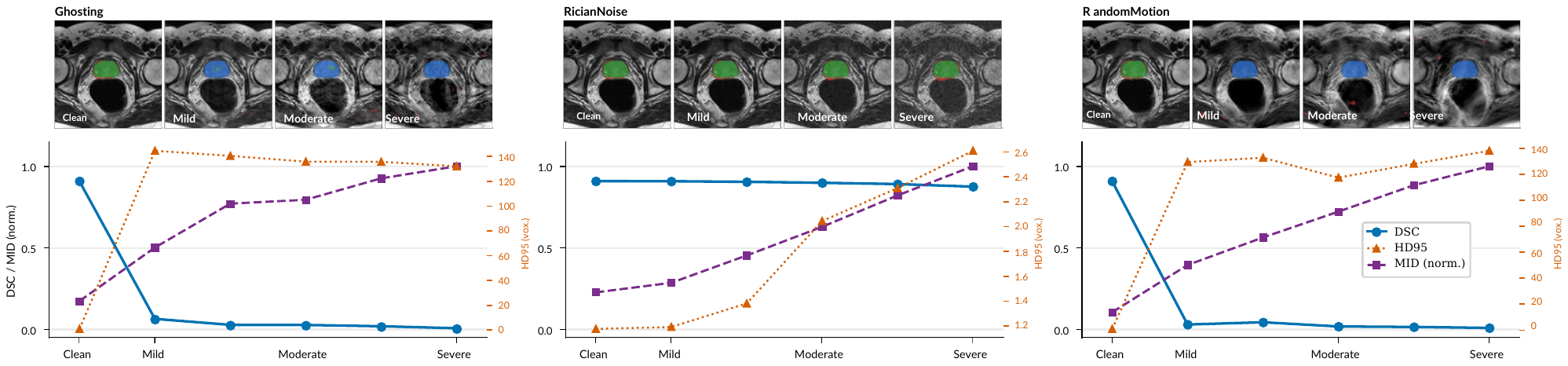}
\caption{ROOD-MRI prostate on Task05\_Prostate using SegResNet-300. For each artefact type, representative T2 slices across severity levels are shown above, with population metrics below.}
\label{fig:rood_prostate}
\end{figure}

\subsection{Metric Comparison}
\label{sec:exp3}

\todo{We compare scalar \MID{} against distributional baselines and per-sample metrics on the same experimental setup as Experiment~1 (\num{56} organ groups $\times$ \num{5} corruption types $\times$ \num{3} severities $= 840$ cells). 
Standard FID applied to Inception-v3 features of axial mid-slice renderings achieves $\rho = -0.38$ with Dice, discarding both volumetric and mask information. Replacing Inception with MedSigLIP mask features in full 3-D yields $\rho = -0.47$, a gain attributable to volumetric encoding. FID-img is essentially zero ($\rho = -0.03$) because images are held fixed across the sweep and only masks are corrupted, confirming that mask information is needed to track segmentation quality. Mahalanobis distance from the joint reference Gaussian reaches $\rho = -0.42$, weaker than \MID{} because it averages per-sample distances rather than comparing the full candidate distribution to the reference. Joint \MID{} reaches $\rho = -0.55$ with Dice, with an absolute gain of \num{0.08} over FID-mask, (overlapping CIs), and $\rho = 0.72$ with HD95 without any per-sample pairing. Full Spearman correlations with \SI{95}{\percent} bootstrap CIs (\num{1000} replicates) for all metrics are in Table~\ref{tab:exp6} (Appendix~\ref{app:corr_figs}). FID-mask, the closest distributional baseline, operates on mask features alone and cannot capture image-mask agreement.}

\subsection{Real Segmentation Models}
\label{sec:exp4}

We evaluate \MID{} on real model predictions in two settings: heterogeneous model tracking against the global reference distribution, and vessel-model ranking on IXI MRA. For the first setting, we use SegResNet at three training stages (SR-50, SR-150, SR-300), frozen SwinUNETR and MedSAM predictions, on Task03 Liver, Task06 Lung, Task07 Pancreas, Task09 Spleen, Task10 Colon, and ACDC. For each model--dataset--foreground-class group, the candidate distribution is built from disjoint evaluation images paired with model predictions and compared against the global reference pool. Dice and HD95 are computed on the same cases as paired-reference quality indicators.

\begin{figure}
\centering
\includegraphics[width=\linewidth]{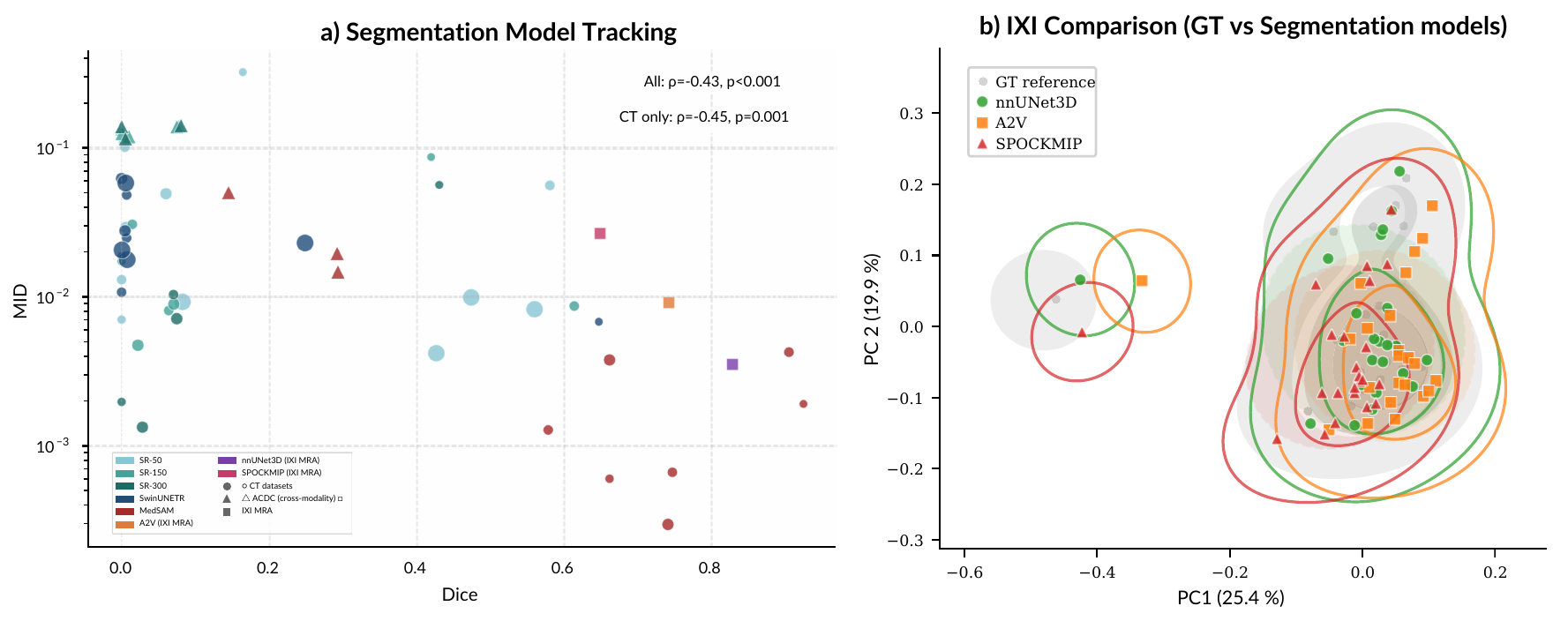}
\caption{\textbf{a)} \MID{} vs.\ Dice across real segmentation model outputs. Points are coloured by model; y-axis is log-scaled. \textbf{b)} Ground-truth and predicted-mask distributions for three IXI MRA vessel segmentation methods.}
\label{fig:model_ranking}
\end{figure}

Figure~\ref{fig:model_ranking}a shows \MID{} against Dice for these groups. On CT datasets, \MID{} decreases as SegResNet training progresses (SR-50 $\to$ SR-150 $\to$ SR-300), and frozen models cluster at lower \MID{} and higher Dice. Across CT groups, \MID{} correlates negatively with Dice (Spearman $\rho=-0.45$); including ACDC gives $\rho=-0.43$ ($p<0.001$). Given that the reference is global, ACDC lies away from the CT trend: its elevated \MID{} reflects cardiac-MRI/reference mismatch rather than segmentation quality alone. For IXI MRA, we use VesselVerse predictions from nnUNet, A2V, and SPOCKMIP~\cite{falcetta2025}. \MID{} ranks the three vessel segmentation methods in the same order as Dice and HD95: nnUNet has the lowest \MID{} ($0.0035$) and highest Dice ($0.83$), followed by A2V ($0.0091$, Dice $0.74$) and SPOCKMIP ($0.026$, Dice $0.65$). The same ordering is observed for HD95 (\SI{4.6}{\mm} vs.\ \SI{20.7}{\mm} for nnUNet and SPOCKMIP), indicating that \MID{} recovers the model ranking without requiring per-case annotations at inference time. Additional setup details are in Appendix~\ref{app:model_ranking}.

\begin{figure}[h]
\centering
\includegraphics[width=0.9\linewidth]{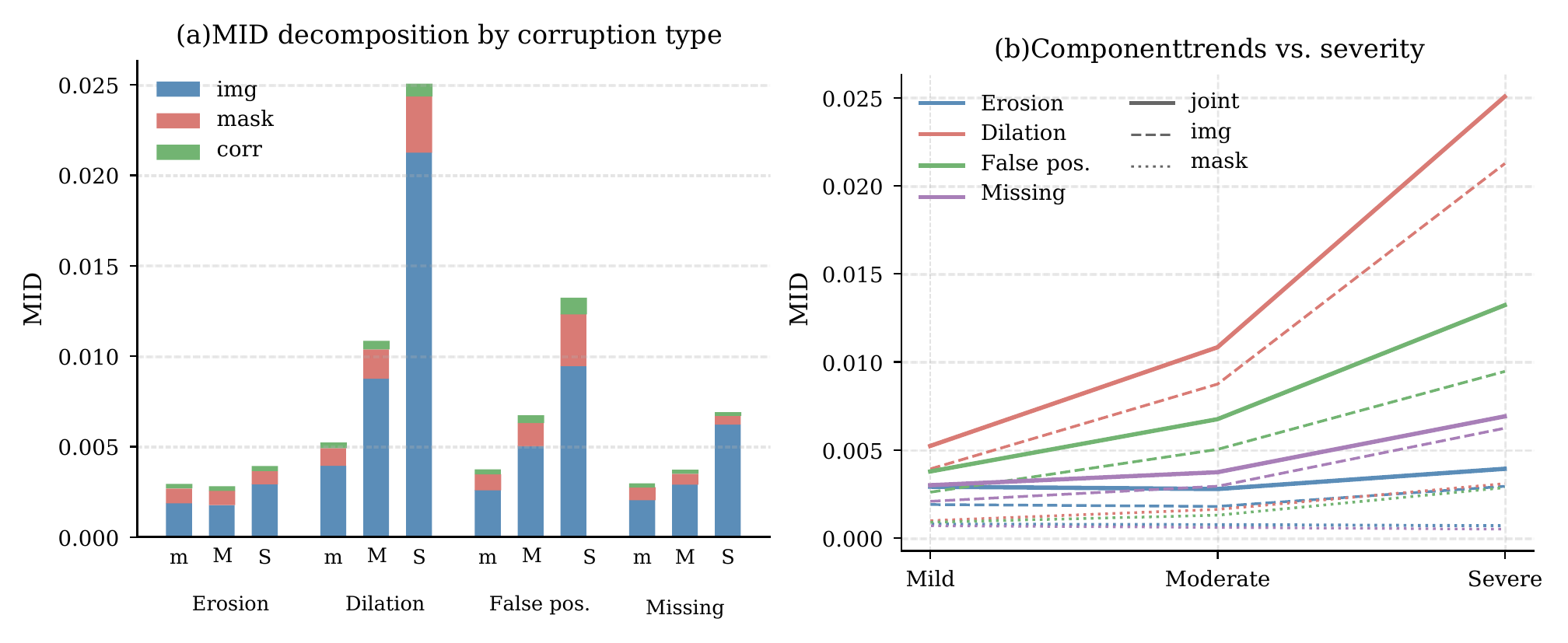}
\caption{\todo{\MID{} decomposition across corruption types on TotalSegmentator v2 CT.
\textbf{(a)} Stacked bars show the image, mask, and residual correspondence terms for each corruption type and severity level (m = mild, M = moderate, S = severe).
\textbf{(b)} Component trends as severity increases.}}
\label{fig:decomposition}
\end{figure}

\subsection{Image-Mask Correspondence}
\label{sec:exp5}
We use the diagnostic decomposition from Section~\ref{sec:decomposition} to test whether \MID{} separates image appearance, mask geometry, and image-mask correspondence. The main correspondence test uses two controlled TotalSegmentator CT settings. Scenario~A corrupts masks while preserving image-mask pairing. Scenario~B randomly shuffles image-mask pairs, preserving image-only and mask-only distributions but disrupting correspondence. Both use the same TotalSegmentator CT reference distribution and $n=300$ feature vectors per organ. Figure~\ref{fig:decomposition} visualises the per-corruption component breakdown for Scenario~A.

Under pair shuffling, $\MID_{\mathrm{img}}$ and $\MID_{\mathrm{mask}}$ remain near zero, while the residual correspondence term carries the signal. This isolates image-mask dependence: the marginal feature distributions are unchanged, but the joint distribution is not. Thus, $\MID_{\mathrm{corr}}$ detects broken image-mask correspondence even when image-only and mask-only statistics are preserved. As an attribution check, Scenario~C measures cross-modality shift between TotalSegmentator CT and AMOS MRI over six shared abdominal organs: spleen, left kidney, right kidney, liver, stomach, and pancreas. Here, $\MID_{\mathrm{img}}$ dominates, consistent with modality-driven image appearance shift. Additional decomposition details are in Appendix~\ref{app:decomposition_extra}.

\section{Discussion}
\label{sec:discussion}

The experiments show that \MID{} provides a consistent distribution-level quality signal across controlled corruptions, real model outputs, and cross-dataset shifts. Its correlation with Dice and HD95 under controlled corruptions ($\rho = -0.55$ and $0.72$) holds without requiring per-sample GT pairing, confirming the metric's practical utility in deployment settings where annotations are unavailable. The graded response to distributional shift, from same-distribution held-out data ($\MID = 0.011$) through modality shift ($0.278$--$0.949$), demonstrates that \MID{} is sensitive to the type of shift and not only its magnitude. 
\todo{The main limitation is the Gaussian approximation: when the goal is within-anatomy monitoring, anatomy-specific references may be preferable. The anatomy-dependence of the absolute scale is a related consequence: \MID{} values for the liver and for the adrenal gland are not directly comparable, and the metric should be interpreted against anatomy-specific reference ranges (Appendix~\ref{app:atlas}).} \MID{} depends on the quality of the underlying encoder. The choice of MedSigLIP is motivated empirically and by its large-scale medical pretraining, but the framework is encoder-agnostic: as stronger medical vision encoders become available, \MID{} can adopt them without changes to the metric definition or estimation procedure. Computationally, feature extraction is the bottleneck. Once features are pre-extracted (approximately 11 seconds per volume on a single GPU), fitting PCA and computing \MID{} takes under one minute on a CPU node for the full \todo{\num{11223}-vector} reference set, making the metric practical for routine evaluation. A conditional variant that conditions the reference distribution on anatomy class would allow cross-anatomy comparison without restricting the reference set. Extending the validation to inherently 2-D modalities (fundus photography, histopathology, X-ray) is a natural next step.

\section{Conclusion}
\label{sec:conclusion}

We have introduced \MID{}, a distribution-level evaluation metric for medical image segmentation. \MID{} embeds image-mask pairs jointly through a pretrained medical vision encoder, fits Gaussian approximations of the reference and candidate distributions, and reports their squared 2-Wasserstein distance. The metric is non-negative, symmetric, and consistent; a diagnostic decomposition into image, mask, and residual correspondence terms is available for failure-mode analysis. Across 13 datasets spanning CT, MRI, and MRA, \MID{} tracks per-sample geometric metrics under controlled corruptions, detects cross-dataset and cross-modality distributional shifts, and recovers correct model rankings without per-case annotations at inference time.

\section{Acknowledgements}
This work is funded by the European Union (ERC CARAVEL, 101171357) and by the French government, through the 3IA Côte d’Azur Investments project (ANR-23-IACL-0001).  Views and opinions expressed are those of the authors only and do not necessarily reflect those of the EU or the European Research Council. Neither the EU nor the granting authority can be held responsible for them.

\bibliographystyle{plain}
\bibliography{references}
\newpage

\appendix

\section{Dataset and Implementation Details}
\label{app:datasets}
This appendix summarises the datasets and implementation details used throughout the experiments. Table~\ref{tab:datasets} reports the modality, number of foreground structures, and number of pre-extracted per-class image-mask feature representations for each dataset.

Feature representations are extracted once with a frozen MedSigLIP encoder and reused for \MID{} and all distributional baselines. For each image-mask pair, the image embedding is computed once and shared across foreground classes, whereas each class-specific binary mask is embedded separately. Thus, $N_{\text{repr}}$ denotes the number of per-class image-mask representations, not the number of subjects. Unless otherwise stated, PCA is fitted on the reference split with $k=128$ components, and covariance matrices are estimated using Ledoit--Wolf shrinkage. \todo{The $56$ foreground classes used in the main corruption sweep (Section~\ref{sec:exp1}) are the subset for which all five corruption types produced valid non-empty masks across the evaluation split; the explicit list of these classes is provided in the released metadata accompanying the code.}
\begin{table}[ht!]
\centering
\caption{Datasets used in the experiments. \todo{Multimodal datasets are listed once per modality, giving \num{13} entries in total. Datasets marked $^\bullet$ are held out from the reference pool and used only as candidate sets for cross-dataset evaluation; the remaining \num{10} datasets form the reference pool described in Section~\ref{sec:setup}.} $^*$56 foreground classes are used across CT and MRI; the full 104-class annotation is available. $^\ddagger$Peripheral zone (PZ) and transition zone (TZ).}
\label{tab:datasets}
\small
\begin{tabular}{lccc}
\toprule
Dataset & Modality & Structures & $N_\text{repr}$ \\
\midrule
\todo{AbdomenCT-1K$^\bullet$~\cite{ma2021abdomenct}}              & CT  & $4$   & $2888$ \\
ACDC~\cite{bernard2018deep}                                  & MRI & $3$   & $900$ \\
\todo{AMOS\,CT$^\bullet$~\cite{ji2022amos}}                       & CT  & \todo{$15$}  & \todo{$3000$} \\
\todo{AMOS\,MRI$^\bullet$}                                       & MRI & \todo{$13$}  & \todo{$520$} \\
IXI MRA (VesselVerse~\cite{ixi2006dataset,falcetta2025})      & MRA & $1$   & $49$ \\
Task03 Liver (Decathlon~\cite{antonelli2022medical})          & CT  & $2$   & $262$ \\
Task05 Prostate (Decathlon)                                  & MRI & $2^\ddagger$ & $64$ \\
Task06 Lung (Decathlon)                                      & CT  & $1$   & $63$ \\
Task07 Pancreas (Decathlon)                                  & CT  & $2$   & $562$ \\
Task09 Spleen (Decathlon)                                    & CT  & $1$   & $41$ \\
Task10 Colon (Decathlon)                                     & CT  & $1$   & $126$ \\
TotalSegmentator v2 CT~\cite{wasserthal2023totalsegmentator} & CT  & $104^*$ & $7368$ \\
TotalSegmentator v2 MRI                                      & MRI & $104^*$ & $1788$ \\
\midrule
Total & & & \todo{$17631$} \\
\bottomrule
\end{tabular}
\end{table}

\section{Controlled Corruption Protocol}
\label{app:corruptions}

This appendix describes the controlled mask-corruption protocol used in Experiment~1. The goal is to create candidate segmentation distributions with known degradation type and severity while keeping the image distribution fixed. This allows us to test whether \MID{} responds monotonically to segmentation quality degradation without using per-sample pairing during metric computation.

We apply five signed-distance-transform (SDT)-based corruption types: elastic boundary deformation, smooth erosion, smooth dilation, blob-shaped false positives, and partial missing slices. Each corruption is applied at three severity levels: mild, moderate, and severe. These corruptions are designed to mimic common segmentation model failures, including boundary bias, over-segmentation, false-positive regions, and incomplete volumetric predictions.

\begin{figure}[h!]
\centering
\includegraphics[width=\linewidth]{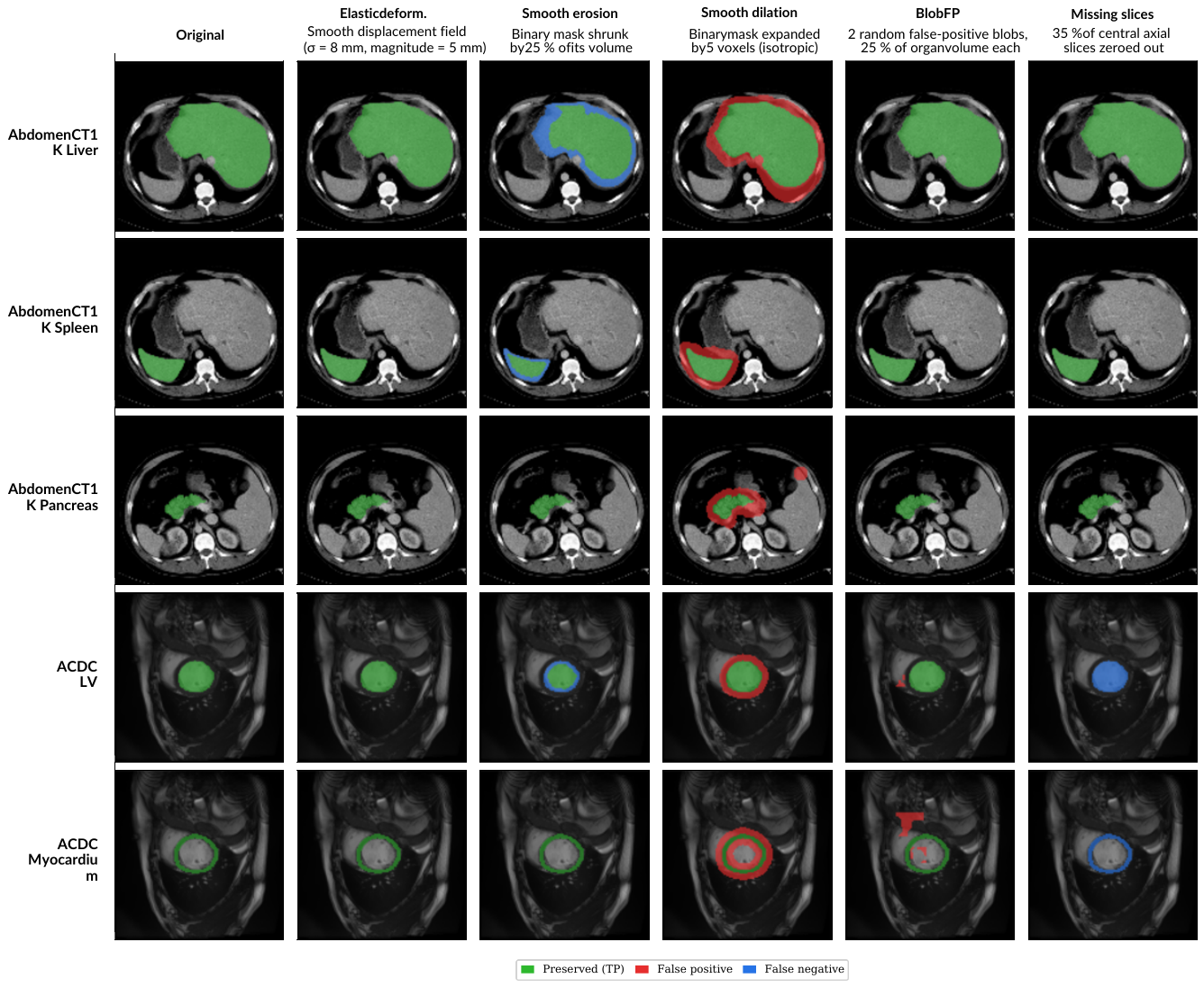}
\caption{Anatomically realistic SDT-based mask corruptions applied to five organs. Columns show the original mask and five corruption types: elastic deformation, smooth erosion, smooth dilation, blob-shaped false positives, and partial missing slices. Overlay colours: green = preserved region, red = false positive, blue = false negative.}
\label{fig:corruptions}
\end{figure}

Figure~\ref{fig:corruptions} illustrates the five corruption types on representative organs. Green denotes preserved true-positive regions, red denotes false positives, and blue denotes false negatives.

\section{Additional Results for Controlled Corruptions}
\label{app:controlled_extra}

This appendix provides additional analyses for Experiment~1. The main text reports the overall correlation between \MID{} and paired-reference metrics under controlled mask corruptions. Here we further examine clean-reference variability, anatomy-specific calibration ranges, and the limitation of comparing absolute \MID{} values across anatomies.

\subsection{Clean-Reference Variability}
\label{app:exp1_groups}

We first estimate the natural variability of \MID{} when both reference and candidate samples are clean. For each dataset--organ group, we split the clean ground-truth image-mask pairs into two disjoint subsets and compute \MID{} between them. This clean-vs-clean setting provides a null baseline for interpreting non-zero \MID{} values caused by finite-sample estimation and anatomical variability.
\begin{figure}[h!]
\centering
\includegraphics[width=0.55\linewidth]{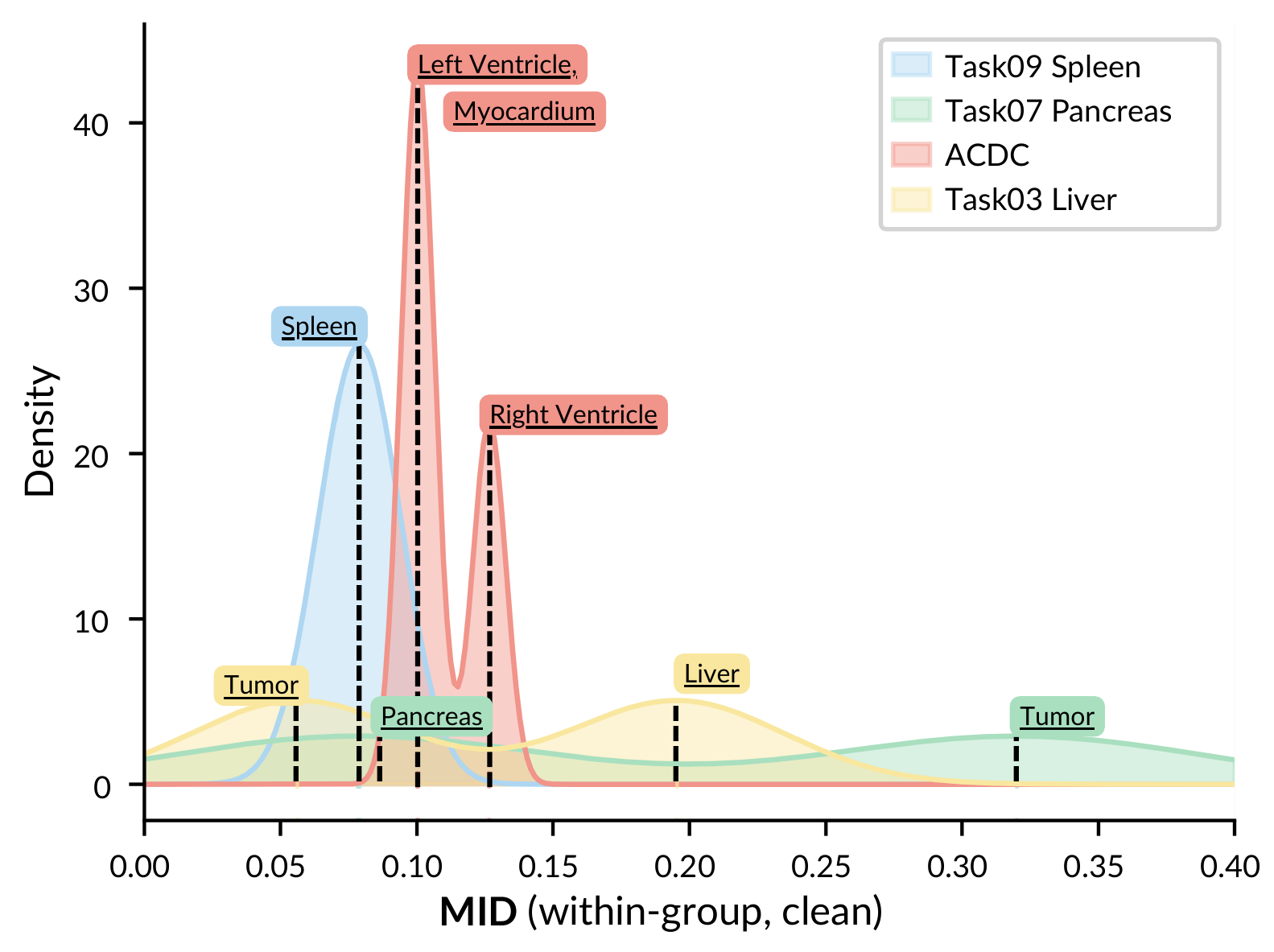}
\caption{Within-group \MID{} distribution across datasets. Each curve shows the kernel density estimate of clean-vs-clean \MID{} values over organ groups within a dataset.}
\label{fig:within_spread_kde}
\end{figure}

\begin{figure}[h!]
\centering
\includegraphics[width=0.9\linewidth]{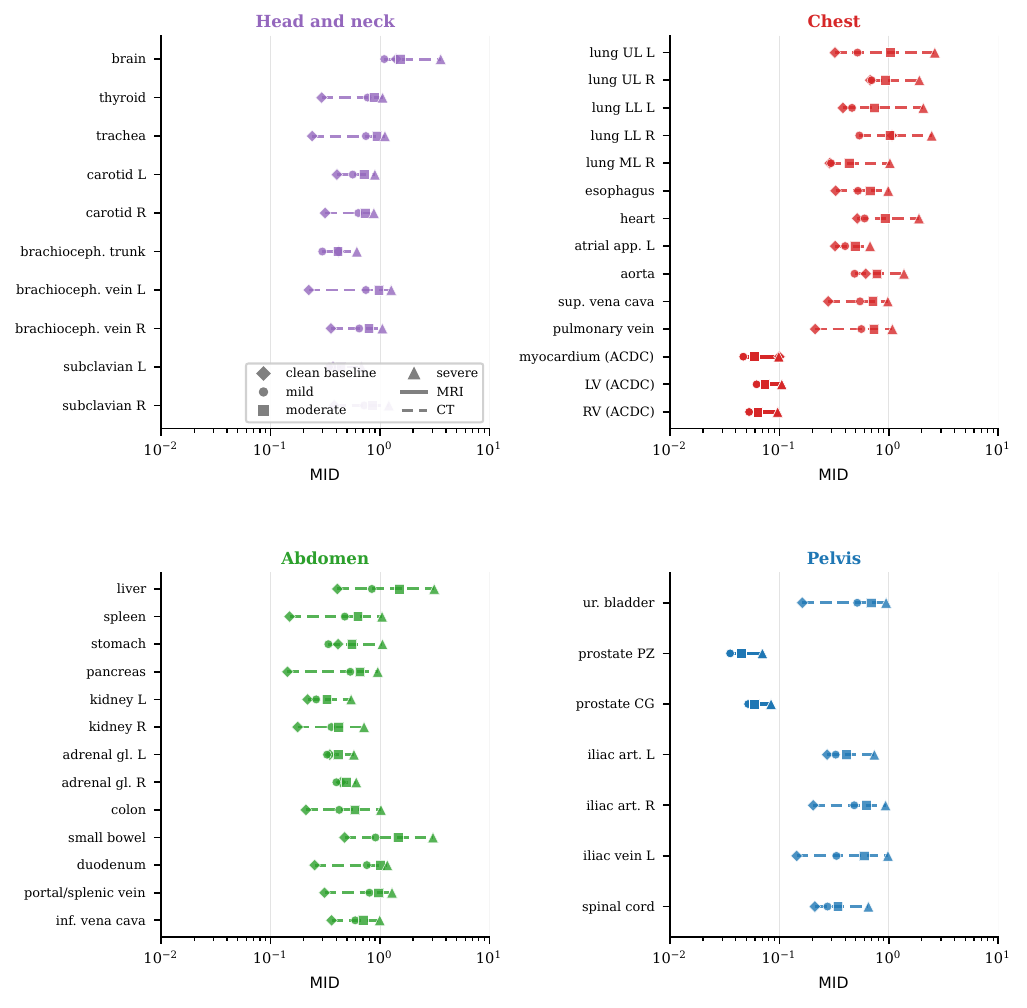}
\caption{\todo{Per-anatomy \MID{} reference index across the organ groups: CT/MRI structures from TotalSegmentator and overlapping MSD/AbdomenCT-1K annotations, plus cerebral arteries from IXI MRA. Each horizontal bar spans the within-group clean-baseline \MID{} (diamond) to the severe-corruption \MID{} (triangle), with mild (circle) and moderate (square) markers at intermediate levels. Values are averaged over five SDT-based corruption types. Solid lines indicate MRI-derived organs; dotted lines indicate CT organs.}}
\label{fig:atlas}
\end{figure}
Figure~\ref{fig:within_spread_kde} reports \MID{} for clean-vs-clean comparisons within each dataset--organ group ($n=56$ per group, with reference and candidate sets drawn from disjoint halves of the clean feature files). Dice is 1.0 for all groups by construction. The observed spread reflects natural anatomical variability and finite-sample estimation noise. This confirms that \MID{} should be interpreted relative to anatomy-specific reference ranges rather than by a universal absolute threshold.

\subsection{Per-Anatomy Reference Index}
\label{app:atlas}

\todo{Because \MID{} is computed in a learned feature space, its absolute scale depends on the anatomy being evaluated. Large organs and compact structures have different baseline variability and different responses to the same corruption type. Figure~\ref{fig:atlas} therefore provides a per-anatomy reference index, showing clean-baseline and corrupted-reference ranges for each foreground class.}

\todo{Overall, the per-anatomy reference index shows that absolute \MID{} values can vary by roughly an order of magnitude across anatomies. This supports the recommendation in Section~\ref{sec:discussion}: \MID{} should be interpreted within an anatomy or relative to an anatomy-specific reference range.}

\section{Distributional Shift and ROOD-MRI Details}
\label{app:shift_rood}

This appendix provides additional details for the distributional-shift experiments in Section~\ref{sec:exp2}. We consider two complementary settings: cross-dataset shifts, where anatomy or modality differs from the reference population, and within-dataset image-quality shifts induced by MRI artefacts. These analyses test whether \MID{} responds to changes in the joint image-mask distribution beyond synthetic mask corruptions.

\subsection{Cross-Dataset Shift Setup}

For the cross-dataset experiment, the reference distribution is built from TotalSegmentator CT training cases over six abdominal organs shared across candidate datasets. Candidate distributions are constructed from held-out TotalSegmentator CT, TotalSegmentator MRI, Task05 Prostate MRI, and ACDC cardiac MRI. This design creates an ordered sequence of shifts: same dataset and modality, modality shift, and combined anatomy--modality shift.

The in-distribution held-out TotalSegmentator CT split provides the finite-sample floor for \MID{}. Candidate datasets with different modality or anatomy are expected to produce progressively larger scores. This setup is used in Table~\ref{tab:ood} in the main text.

\subsection{ROOD-MRI Prostate Robustness}
\label{app:rood_prostate}

This subsection gives the full setup for the within-dataset artefact experiment. The reference distribution is computed once from clean Task05\_Prostate T2 images paired with ground-truth masks. For each artefact type and severity level, the candidate distribution is built from corrupted T2 images paired with SegResNet predictions on those corrupted images. Ground-truth masks are used only to compute Dice and HD95 for analysis, not to compute \MID{} at inference time.

\paragraph{Setup.}
Dataset: Task05\_Prostate~\cite{antonelli2022medical}, consisting of T2-weighted prostate MRI with peripheral-zone and transition-zone labels. Model: SegResNet trained for 300 epochs on Task05\_Prostate~\cite{myronenko20183d}. Artefacts: Ghosting, Rician noise, and Random Motion at three severity levels: mild, moderate, and severe. Each artefact is applied to the T2 channel only; the ADC channel is kept clean.

\paragraph{\MID{} at inference time.}
The reference distribution is computed offline from clean T2 images paired with ground-truth masks. For each artefact--severity condition, the candidate distribution is formed using corrupted T2 images paired with SegResNet predictions. Thus, \MID{} does not use a ground-truth mask for each candidate prediction. Dice and HD95, by contrast, require paired ground-truth masks and are used only as reference-quality indicators.

\paragraph{Interpretation.}
As reported in Section~\ref{sec:exp2}, \MID{} increases with artefact severity and follows the degradation trends measured by Dice and HD95. This supports its use for monitoring distribution-level segmentation quality when per-case annotations are unavailable.

\section{Metric Comparison: Full Correlation Results}
\label{app:corr_figs}

\todo{This appendix reports the full Spearman correlations with \SI{95}{\percent} bootstrap confidence intervals for the metric comparison in Section~\ref{sec:exp3}. Each distributional metric is correlated against \emph{both} per-sample paired metrics, Dice and HD95, computed cell-by-cell on the same $840$ cells. The comparison includes joint \MID{} and five feature-space distributional baselines: FID on Inception-v3 mid-slice features, FID on MedSigLIP image features, FID on MedSigLIP mask features, KID on MedSigLIP mask features, and Mahalanobis distance from the joint reference Gaussian. Dice and HD95 themselves are not listed as rows: they are the reference targets that the columns are computed against. The decomposition of \MID{} into $\MID_{\mathrm{img}}$, $\MID_{\mathrm{mask}}$, and $\MID_{\mathrm{corr}}$ is examined separately in Experiment~5 and Appendix~\ref{app:decomposition_extra} and is not included here.}

\todo{\begin{table}[h]
\centering
\caption{Metric comparison on the controlled corruption sweep. Spearman $\rho$ is computed cell-by-cell against mean Dice and against HD95 across $840$ cells ($56$ organ groups $\times$ 5 corruption types $\times$ 3 severity levels); bootstrap CIs use 1000 replicates. The two reference targets are inversely related (Dice goes up, HD95 goes down for better segmentations), so a metric that tracks segmentation quality should produce a negative $\rho$ against Dice and a positive $\rho$ against HD95 of comparable magnitude.}
\label{tab:exp6}
\small
\setlength{\tabcolsep}{4.5pt}
\renewcommand{\arraystretch}{1.12}
\begin{tabular}{llccccc}
\toprule
\multirow{2}{*}{Category} & \multirow{2}{*}{Metric} & \multirow{2}{*}{Encoding}
 & \multicolumn{2}{c}{vs.\ Dice} & \multicolumn{2}{c}{vs.\ HD95} \\
\cmidrule(lr){4-5} \cmidrule(lr){6-7}
 &  &  & $\rho$ & \SI{95}{\percent} CI & $\rho$ & \SI{95}{\percent} CI \\
\midrule
\textit{Ours}
 & \MID{}                 & MedSigLIP    & $-0.55$ & $[-0.61,\,-0.49]$ & $+0.72$ & $[+0.66,\,+0.77]$ \\
\midrule
\multirow{5}{*}{Baselines}
 & FID-mask, 3-D          & MedSigLIP    & $-0.47$ & $[-0.53,\,-0.40]$ & $+0.61$ & $[+0.54,\,+0.67]$ \\
 & FID-img, 3-D           & MedSigLIP    & $-0.03$ & $[-0.10,\,+0.04]$ & $+0.04$ & $[-0.04,\,+0.12]$ \\
 & KID                    & MedSigLIP    & $-0.28$ & $[-0.36,\,-0.20]$ & $+0.36$ & $[+0.28,\,+0.44]$ \\
 & FID-Inception, 2-D     & Inception-v3 & $-0.38$ & $[-0.56,\,-0.18]$ & $+0.49$ & $[+0.31,\,+0.65]$ \\
 & Mahalanobis (joint)    & MedSigLIP    & $-0.42$ & $[-0.49,\,-0.34]$ & $+0.55$ & $[+0.47,\,+0.62]$ \\
\bottomrule
\end{tabular}
\end{table}}

\todo{Two patterns are visible. First, \MID{} is the strongest correlator with both Dice ($\rho = -0.55$) and HD95 ($\rho = +0.72$) among the unpaired distributional metrics; the gap to FID-mask is consistent across the two reference targets ($+0.08$ absolute against Dice, $+0.11$ absolute against HD95), and the bootstrap CIs of MID and FID-mask overlap in both columns. Second, the HD95 column has uniformly larger $|\rho|$ than the Dice column for every metric. This is consistent with HD95 being more sensitive to the boundary-shape errors that dominate the corruption sweep (smooth erosion, dilation, elastic deformation) while Dice penalises overlap loss more uniformly. FID-img is essentially zero in both columns, as expected when images are held fixed across the sweep.}

\section{Image-Mask Correspondence Decomposition}
\label{app:decomposition_extra}

This appendix gives the formal block-covariance derivation that underlies the diagnostic decomposition used in Section~\ref{sec:decomposition}, and then reports the empirical setup that follows from it.

\subsection{Covariance Block Structure}
\label{app:cov_block}

\todo{For an image-mask pair $(X, m_c)$ and the per-class joint embedding of Definition~\ref{def:joint}, write the projected joint vector as $z = (z_x; z_m) \in \R^{k}$. For a distribution $P$ over such embeddings, the mean and covariance admit the block form}
\begin{equation}
\todo{\mu_P = \begin{pmatrix} \mu^x_P \\ \mu^m_P \end{pmatrix},
\qquad
\Sigma_P = \begin{pmatrix} \Sigma^{xx}_P & \Sigma^{xm}_P \\ \Sigma^{mx}_P & \Sigma^{mm}_P \end{pmatrix},}
\label{eq:block_cov}
\end{equation}
\todo{with $\Sigma^{xm}_P = (\Sigma^{mx}_P)^{\top}$, and analogously for $Q$. The squared 2-Wasserstein distance between Gaussian fits $\mathcal{N}(\mu_P, \Sigma_P)$ and $\mathcal{N}(\mu_Q, \Sigma_Q)$ is}
\begin{equation}
\todo{d_F^2(P, Q) = \|\mu_P - \mu_Q\|^2 + \mathrm{Tr}\!\left(\Sigma_P + \Sigma_Q - 2\bigl(\Sigma_P^{1/2}\Sigma_Q\Sigma_P^{1/2}\bigr)^{1/2}\right).}
\label{eq:fid_full}
\end{equation}
\todo{Two of the three terms in (\ref{eq:fid_full}) are linear in the block partition. The mean term decomposes as}
\begin{equation}
\todo{\|\mu_P - \mu_Q\|^2 = \|\mu^x_P - \mu^x_Q\|^2 + \|\mu^m_P - \mu^m_Q\|^2,}
\label{eq:mean_split}
\end{equation}
\todo{and the additive trace term decomposes as}
\begin{equation}
\todo{\mathrm{Tr}(\Sigma_P + \Sigma_Q) = \mathrm{Tr}(\Sigma^{xx}_P + \Sigma^{xx}_Q) + \mathrm{Tr}(\Sigma^{mm}_P + \Sigma^{mm}_Q).}
\label{eq:trace_split}
\end{equation}
\todo{The geometric-mean term $\mathrm{Tr}\bigl(\Sigma_P^{1/2}\Sigma_Q\Sigma_P^{1/2}\bigr)^{1/2}$ does \emph{not} decompose linearly in general, because the principal matrix square root mixes diagonal and off-diagonal blocks whenever either $\Sigma^{xm}_P$ or $\Sigma^{xm}_Q$ is non-zero.}

\subsection{Image, Mask, and Residual Correspondence Terms}

\todo{Following the FJD-style construction~\cite{devries2019evaluation}, we define the marginal Fréchet distances on the image and mask blocks alone:}
\begin{align}
\todo{\MID_{\mathrm{img}}(P, Q)}  &\todo{= \|\mu^x_P - \mu^x_Q\|^2 + \mathrm{Tr}\!\left(\Sigma^{xx}_P + \Sigma^{xx}_Q - 2\bigl(\Sigma^{xx,1/2}_P \Sigma^{xx}_Q \Sigma^{xx,1/2}_P\bigr)^{1/2}\right),} \label{eq:mid_img} \\
\todo{\MID_{\mathrm{mask}}(P, Q)} &\todo{= \|\mu^m_P - \mu^m_Q\|^2 + \mathrm{Tr}\!\left(\Sigma^{mm}_P + \Sigma^{mm}_Q - 2\bigl(\Sigma^{mm,1/2}_P \Sigma^{mm}_Q \Sigma^{mm,1/2}_P\bigr)^{1/2}\right).} \label{eq:mid_mask}
\end{align}
\todo{The residual correspondence term is defined by completion to the full joint score:}
\begin{equation}
\todo{\MID_{\mathrm{corr}}(P, Q) = \MID(P, Q) - \MID_{\mathrm{img}}(P, Q) - \MID_{\mathrm{mask}}(P, Q).}
\label{eq:mid_corr}
\end{equation}
\todo{This residual is the contribution of the geometric-mean term in (\ref{eq:fid_full}) that is not captured by either marginal Fréchet distance, and therefore depends only on the off-diagonal blocks $\Sigma^{xm}_P, \Sigma^{xm}_Q$ that quantify image--mask coupling under each distribution.}

\paragraph{Block-diagonal case.}
\todo{If $\Sigma^{xm}_P = 0$ and $\Sigma^{xm}_Q = 0$, then both $\Sigma_P$ and $\Sigma_Q$ are block-diagonal, $\Sigma_P^{1/2}\Sigma_Q\Sigma_P^{1/2}$ is block-diagonal with blocks $\Sigma^{xx,1/2}_P\Sigma^{xx}_Q\Sigma^{xx,1/2}_P$ and $\Sigma^{mm,1/2}_P\Sigma^{mm}_Q\Sigma^{mm,1/2}_P$, the principal square root acts block-wise on positive semi-definite block-diagonal matrices, and the trace decomposes per block. Substituting into (\ref{eq:fid_full}) and using (\ref{eq:mean_split})--(\ref{eq:trace_split}) gives}
\begin{equation}
\todo{\Sigma^{xm}_P = \Sigma^{xm}_Q = 0 \;\;\Longrightarrow\;\; \MID(P, Q) = \MID_{\mathrm{img}}(P, Q) + \MID_{\mathrm{mask}}(P, Q),}
\end{equation}
\todo{so $\MID_{\mathrm{corr}}(P, Q) = 0$. Conversely, when at least one of the off-diagonal blocks is non-zero and they differ between $P$ and $Q$, the geometric-mean term picks up cross-block contributions and $\MID_{\mathrm{corr}}$ is generally non-zero. We therefore interpret $\MID_{\mathrm{corr}}$ as the part of segmentation quality that is attributable to image--mask correspondence rather than to image or mask marginals alone.}

\subsection{Experimental Scenarios}

This subsection provides additional details for the decomposition experiment in Section~\ref{sec:exp5}. The purpose is to test whether the residual correspondence term $\MID_{\mathrm{corr}}$ responds to changes in image-mask dependence that are not visible from image-only or mask-only marginal distributions.

Scenario~A corrupts masks while preserving image-mask pairing. Scenario~B randomly shuffles masks across images, preserving the image and mask marginal distributions while disrupting their dependence. Scenario~C compares TotalSegmentator CT to AMOS MRI to illustrate how the decomposition behaves under a dominant modality shift. The decomposition is visualised in Figure~\ref{fig:decomposition} in the main text.

\todo{Pair shuffling (Scenario~B) is the cleanest empirical test of the block-structure derivation above: the candidate mask multiset is identical to the un-shuffled case, so $\Sigma^{mm}_Q$ and $\mu^m_Q$ are exactly preserved; image features are also unchanged, so $\Sigma^{xx}_Q$ and $\mu^x_Q$ are exactly preserved; only the off-diagonal block $\Sigma^{xm}_Q$ is destroyed (driven to approximately zero). Consequently, $\MID_{\mathrm{img}}$ and $\MID_{\mathrm{mask}}$ should be invariant under shuffling, while $\MID_{\mathrm{corr}}$ should absorb the entire effect. This is the prediction that Section~\ref{sec:exp5} tests.}

Overall, the experiment supports interpreting $\MID_{\mathrm{corr}}$ as a diagnostic residual for correspondence changes rather than as a standalone metric.

\section{Real-Model Ranking Details}
\label{app:model_ranking}

This appendix provides additional details for the real segmentation model experiment in Section~\ref{sec:exp4}. The goal is to test whether \MID{} ranks real model outputs consistently with paired-reference metrics, rather than only responding to synthetic corruptions.

\todo{For each model--dataset--organ group, the candidate distribution is built from model predictions on disjoint evaluation images, with each prediction paired with the corresponding image latent to form an image-mask embedding. \MID{} is then computed against the global reference pool defined in Section~\ref{sec:setup} (the same 10-dataset reference used throughout the paper), without sample-wise matching between reference and candidate. Using the global reference pool, rather than a group-specific reference, lets us compare candidates from different models, datasets, and anatomies on a common scale; the ACDC offset visible in Figure~\ref{fig:model_ranking}a is a direct consequence of this choice and reflects the CT/MRI mismatch with the predominantly CT reference rather than poor cardiac segmentation quality.}

For the IXI MRA vessel segmentation task, \MID{} ranks the released methods in the same order as Dice and HD95: nnUNet obtains the lowest \MID{} and highest Dice, followed by A2V and SPOCKMIP. This setting is particularly informative because all methods are evaluated on the same anatomy and modality, avoiding cross-modality reference mismatch.

\section{Projection and Sample-Size Analyses}
\label{app:projection_sample}

This appendix analyses two estimator choices used by \MID{}: the PCA projection and the candidate sample size. Since \MID{} relies on Gaussian statistics in a projected feature space, both the projection subspace and the number of candidate samples can affect the stability of the estimate.

\subsection{Dataset-Specific Projections}
\label{app:exp8}

A practical question concerns whether the projection should be fitted on a large universal corpus or on the target dataset alone. The default \MID{} projection is fitted on the full MedSigLIP feature corpus, providing a universal latent coordinate system. A practitioner with access to a single target dataset may instead fit the projection on that dataset, trading cross-dataset comparability for within-domain sensitivity.

We test this per-organ, using TotalSegmentator CT as the healthy reference for two organs that also appear with pathological annotations in MSD: liver (class 1 in Task03, where class 2 is liver tumour) and pancreas (class 1 in Task07, where class 2 is pancreas tumour). For each organ, we fit a $k=128$ PCA projection on the TotalSegmentator CT training split and compare against the universal projection.

\paragraph{Per-organ results.}
Table~\ref{tab:exp8_matrix} reports \MID{} under each projection for three candidate sets: a held-out split from the same TotalSegmentator distribution, the MSD healthy organ class from a different scanner and protocol, and the corresponding tumour class.

\begin{table}[H]
\centering
\caption{\MID{} between TotalSegmentator CT reference and three candidate sets, under universal and organ-specific projections ($k=128$). Reference: TotalSegmentator CT training split; organ-specific PCA is fitted on the same reference.}
\label{tab:exp8_matrix}
\small
\setlength{\tabcolsep}{6pt}
\renewcommand{\arraystretch}{1.1}
\begin{tabular}{llrr}
\toprule
Organ & Candidate & Universal & Organ-PCA \\
\midrule
\multirow{3}{*}{Liver}
  & TotalSeg CT held-out (in-distribution) & 0.050 & 0.049 \\
  & Task03 liver, class~1 (healthy, cross-dataset) & 0.154 & 0.152 \\
  & Task03 liver tumour, class~2 (pathological)    & 0.165 & 0.162 \\
\midrule
\multirow{3}{*}{Pancreas}
  & TotalSeg CT held-out (in-distribution) & 0.038 & 0.037 \\
  & Task07 pancreas, class~1 (healthy, cross-dataset) & 0.238 & 0.236 \\
  & Task07 pancreas tumour, class~2 (pathological)    & 0.216 & 0.214 \\
\bottomrule
\end{tabular}
\end{table}

The two projections produce near-identical values: organ-specific PCA reduces \MID{} by at most 1.5\% relative to the universal projection. In-distribution held-out splits yield the smallest scores in both cases. For liver, the tumour class scores above the healthy cross-dataset class, reflecting the additional distributional shift introduced by pathological appearance; for pancreas, the ordering is reversed, because pancreatic tumour masks in MSD are smaller and more variable than the full-organ masks used for the reference, and this scale difference reduces rather than increases the measured divergence. The practical implication is that the projection choice has negligible effect on relative rankings for single-organ evaluations.

\paragraph{Practical recommendation.}
For single-dataset evaluation, fitting PCA on a held-out reference split of the target data is appropriate and can improve within-domain sensitivity. For cross-dataset comparisons, the universal projection is preferable because it keeps all datasets in a common coordinate system.

\subsection{Sample-Size Sensitivity}
\label{app:sample_complexity}

We next study how \MID{} varies with the candidate set size. This analysis is important because medical imaging datasets often contain substantially fewer cases than natural-image benchmarks used for FID evaluation.

\todo{Table~\ref{tab:sample_complexity} reports the mean and coefficient of variation (CV) of \MID{} as a function of candidate set size $n$, using 50 bootstrap replicates per $n$ in a representative cross-dataset setting (TotalSeg CT reference vs.\ TotalSeg MRI candidate, asymptotic \MID{} $\approx 0.30$, see Table~\ref{tab:ood}). The reference set is fixed at $n=500$.}

\begin{table}[H]
\centering
\caption{\todo{Sample-size sensitivity of \MID{} in a cross-dataset setting (TotalSeg CT reference vs.\ TotalSeg MRI candidate; asymptotic \MID{} $\approx 0.30$). The reference set is fixed at $n=500$; candidate sets are subsampled at different sizes. Mean and coefficient of variation (CV) are computed over 50 bootstrap replicates. The decreasing mean reflects the well-known finite-sample positive bias of the squared 2-Wasserstein distance between Gaussian fits.}}
\label{tab:sample_complexity}
\small
\setlength{\tabcolsep}{6pt}
\renewcommand{\arraystretch}{1.12}
\begin{tabular}{lrrrrrrr}
\toprule
\multirow{2}{*}{Statistic} 
& \multicolumn{7}{c}{Candidate set size $n$} \\
\cmidrule(lr){2-8}
& 10 & 25 & 50 & 100 & 250 & 500 & 1000 \\
\midrule
Mean \MID{} & 1.85 & 0.92 & 0.58 & 0.42 & 0.34 & 0.31 & 0.30 \\
CV          & 32\% & 20\% & 14\% & 10\% & 6.3\% & 4.5\% & 3.2\% \\
\bottomrule
\end{tabular}
\end{table}
The mean estimate decreases monotonically with $n$, consistent with the finite-sample positive bias of the Fréchet estimator, and stabilises around the asymptotic value ($\approx 0.30$) for $n \gtrsim 250$. The CV decreases roughly as $n^{-1/2}$, as expected for bootstrap variability of a moment-based estimator. Comparisons are most reliable when candidate sets have similar sample sizes, or when the evaluation protocol fixes the candidate set size across methods.
\section{Latent Space Geometry}
\label{app:latent_vis}
\label{app:pca_structure}

This appendix visualises the geometry of the MedSigLIP feature space used by \MID{}. These analyses are not used to compute the metric; rather, they provide qualitative evidence that the encoder organizes samples by modality and anatomy in a way that is meaningful for distributional comparison.

\subsection{PCA Projection Structure}

\begin{figure}[H]
\centering
\begin{minipage}[t]{0.43\linewidth}
\centering
\vspace{0pt}
\includegraphics[width=\linewidth]{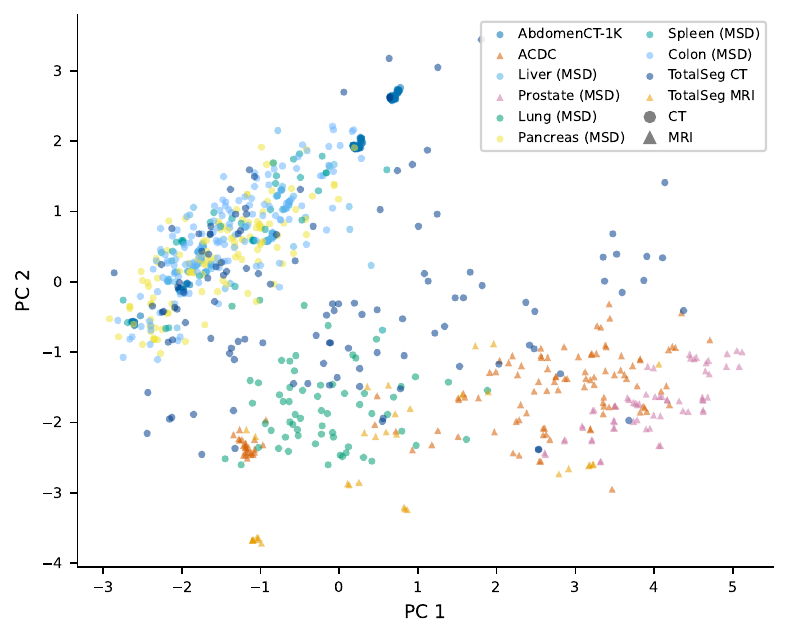}
\captionof{figure}{First two PCA components of joint MedSigLIP embeddings. CT datasets occupy a common region, whereas MRI datasets are displaced along the dominant modality-related axis.}
\label{fig:pca_scatter}
\end{minipage}
\hfill
\begin{minipage}[t]{0.53\linewidth}
\centering
\vspace{0pt}
\captionof{table}{Mean \MID{} between clean and corrupted latents, averaged over organ groups for each dataset. Larger values indicate greater displacement from the reference distribution in the universal PCA space.}
\label{tab:pca_mid}
\small
\setlength{\tabcolsep}{5pt}
\renewcommand{\arraystretch}{1.08}
\begin{tabular}{llcc}
\toprule
Dataset & Modality & Groups & Mean \MID{} \\
\midrule
TotalSegmentator CT  & CT  & 44 & 0.046 \\
Task10 Colon         & CT  &  1 & 0.077 \\
Task09 Spleen        & CT  &  1 & 0.081 \\
AbdomenCT-1K         & CT  &  4 & 0.093 \\
Task03 Liver         & CT  &  2 & 0.099 \\
Task07 Pancreas      & CT  &  2 & 0.105 \\
Task06 Lung          & CT  &  1 & 0.260 \\
\midrule
ACDC                 & MRI &  3 & 0.072 \\
Task05 Prostate      & MRI &  2 & 0.146 \\
TotalSegmentator MRI & MRI &  2 & 0.895 \\
\bottomrule
\end{tabular}
\end{minipage}
\end{figure}

Figure~\ref{fig:pca_scatter} plots the first two principal components of joint MedSigLIP embeddings. CT datasets cluster in one region of the space, while MRI datasets are displaced along the dominant principal component. This supports the interpretation that modality is a major source of variation in the joint feature space.

The PCA visualisation explains why \MID{} naturally assigns larger scores to cross-modality shifts: MRI samples are displaced from CT samples in the joint feature space. For within-modality evaluation, this modality axis contributes much less to the total score.

\subsection{MedSigLIP Feature Clustering}

\begin{figure}[h!]
\centering
\includegraphics[width=\linewidth]{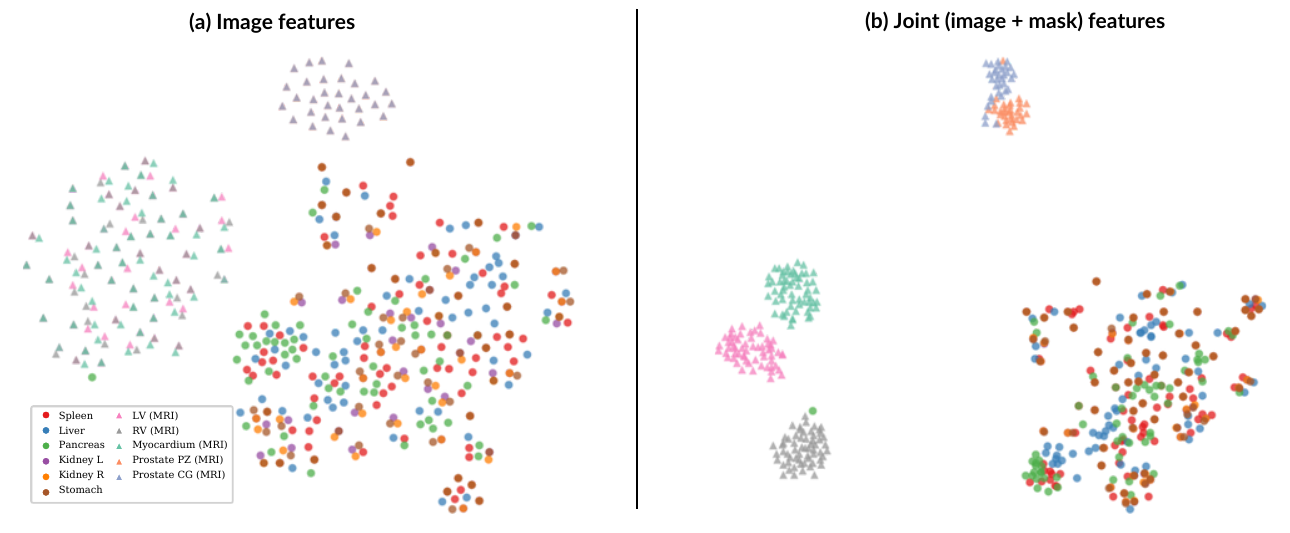}
\caption{t-SNE visualisation of MedSigLIP embeddings for eleven organ types. Left: mask-only embeddings. Right: joint image-mask embeddings. Points are coloured by organ, with circles denoting CT samples and triangles denoting MRI samples. Anatomical structures form coherent clusters in both spaces, whereas MRI samples separate from their CT counterparts mainly in the joint embedding space, suggesting that modality-related variation is primarily captured by the image component.}
\label{fig:latent_overview}
\end{figure}

Figure~\ref{fig:latent_overview} shows t-SNE projections of raw MedSigLIP features before PCA. We visualise both mask-only features and joint image-mask features for representative CT and MRI organs.

\begin{figure}[h!]
\centering
\includegraphics[width=0.85\linewidth]{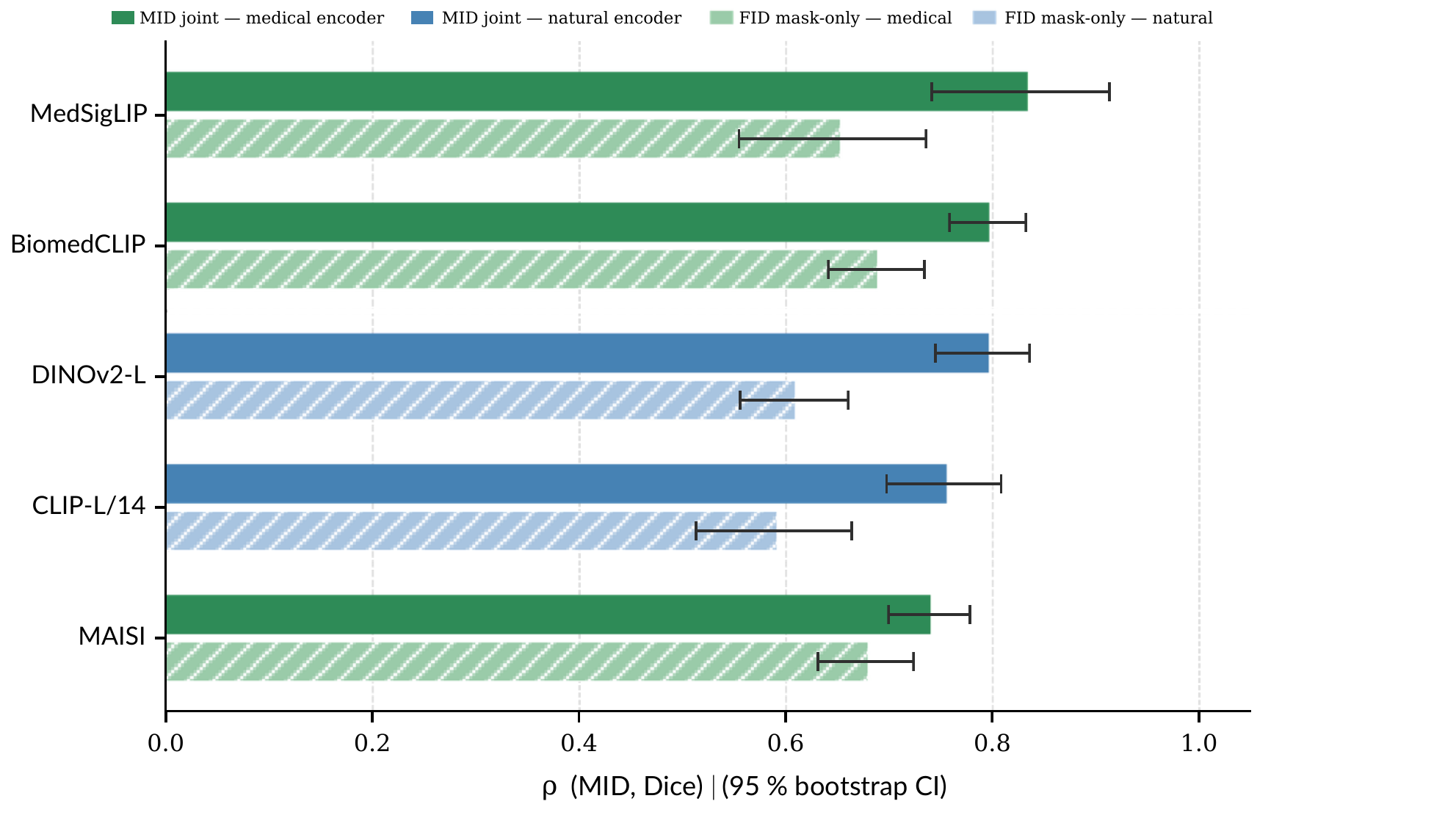}
\caption{Mean per-organ $|\rho(\MID{}, \mathrm{Dice})|$ for five encoder backbones on the TotalSegmentator v2 CT corruption sweep. Each bar reports the mean over per-organ rank correlations.}
\label{fig:encoder_ablation_barplot}
\end{figure}
The clustering supports the use of MedSigLIP as the \MID{} backbone: similar anatomical structures occupy nearby regions of the feature space, while modality shifts are reflected in the joint embedding.

\section{Encoder Backbone Ablation}
\label{app:encoder_choice}

This appendix studies the dependence of \MID{} on the choice of visual encoder. Since \MID{} is an encoder-space distributional metric, its behaviour depends on whether the embedding captures medically meaningful image and mask structure. We compare MedSigLIP against biomedical, natural-image, and generative encoders using the same controlled corruption sweep.

Figure~\ref{fig:encoder_ablation_barplot} summarises the per-organ rank correlation between \MID{} and Dice across five encoder backbones. MedSigLIP achieves the highest mean $|\rho| = 0.833$ $[0.742, 0.918]$; BiomedCLIP and DINOv2-L follow at $0.796$ ($[0.758, 0.833]$ and $[0.750, 0.836]$, respectively), while CLIP-L/14 reaches $0.755$ $[0.698, 0.807]$ and MAISI obtains $0.739$ $[0.699, 0.778]$.

Two patterns emerge. First, joint image-mask encoding outperforms mask-only FID for every encoder, confirming that the image component contributes information beyond mask geometry alone. Second, domain-specific encoders do not uniformly outperform natural-image encoders: DINOv2-L matches BiomedCLIP despite having no medical pretraining, suggesting that representational capacity can partially compensate for domain gap in this corruption setting. The lower correlation for MAISI is consistent with its VAE-style generative objective, which regularises the latent space toward a spherical prior rather than organising it for discriminative anatomy or quality assessment.

\end{document}